# Transfer Portal: Accurately Forecasting the Impact of a Player Transfer in Soccer

Daniel Dinsdale and Joe Gallagher

## 1. Introduction

Deadline day is one of the biggest occasions in the soccer calendar. It is the final opportunity for teams to sign players in the trading window before it is closed for the first half of the season. In 2021, over $200 million was spent in the Premier League on this day alone—including big money signings such as Cristiano Ronaldo to Manchester United—whilst over a full year there has been up to $7.35bn spent globally [1].

As a team owner, manager or transfer committee looking to improve the fortunes of your team on deadline day, these important and time-dependent decisions rely on player scouting to determine potential signings who fit your team's playing style and budget. The scouting process generally combines data appraisal on performance metrics with direct observations of players via video and/or match attendance to make critical business decisions on which players represent best value for money. This is because, in addition to being the most valuable prediction a team makes, it is also the most complex analytics task to perform as consideration is also required for: a) the difference in playing style between the player's current and the target team, b) the difference in teammate ability, c) the difference in league quality and style, and d) the role the player is desired to play. This process requires substantial time investment, which with a rapidly changing market is often not viable or flexible enough to make informed decisions on the fly.

In this paper, we present our *"Transfer Portal"* model, which helps to inform these vital business decisions through the power of deep learning to provide predictions of player performance across 13 player-level metrics.

Our approach enables two key prediction tasks:

1. **Transfer Impact:** Estimate the impact a specific player will have in terms of their player contribution for a proposed future club, and
2. **Player Recommendation:** Using predicted player impact, create a shortlist of players across any number of chosen leagues which represent the best transfer targets, or potential replacements for a departing player.

For example, we can use our approach to estimate the impact that Harry Kane's rumoured move from Tottenham to Manchester City over summer 2021 would have had, or the impact of Messi leaving Barcelona to Paris Saint-Germain. We can also recommend which players both Tottenham and Barcelona could target as replacements. These are the obvious cases, but the real power of our Transfer Portal approach is to highlight key players in teams which do not have the financial backing of top teams.

A good example is Stade Rennais FC (Rennes)—a French Ligue 1 team who have made some great recruitment decisions in recent years, especially on the wing. Examples include: Ismaïla Sarr who they bought from Metz for £15m in 2017 and sold to Watford for £27m two years later; Sarr's replacement Raphinha, who joined from Sporting in 2019 before departing for Leeds the following season; and Raphinha's replacement Jérémy Doku, who came in from Anderlecht for £23m in October 2020 and is already being touted with big money moves to top teams in the Premier League.

Imagine if there was a predictive model which a top-tier team such as Liverpool could use to forecast the impact Doku would have? Additionally, if that same model could be used by Rennes to recommend which player might be a good replacement for Doku? ***Using our Transfer Portal prediction model, this is now possible.*** An example of both these use cases are illustrated in Figure 1 Figure 2—with Figure 1 showing the predicted performance Doku would have at Liverpool, and Figure 2 showing how we can use our model to recommend and assess candidate players to replace Doku at Stade Rennais FC.

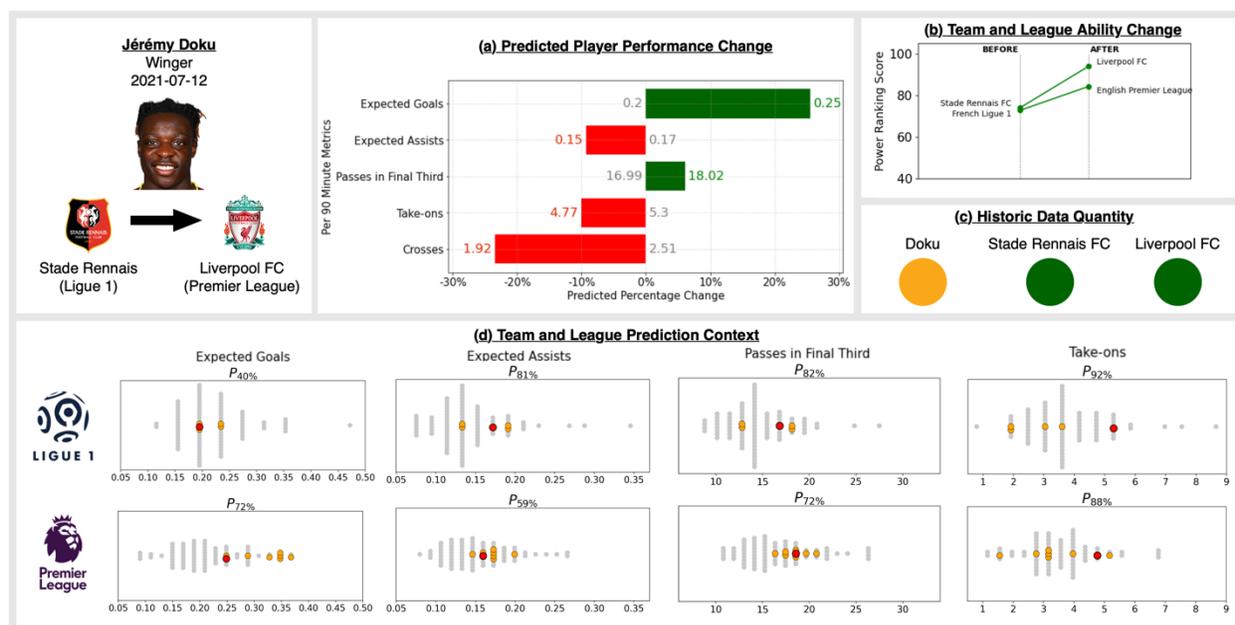

*Figure 1: (a) First we predict the change in individual players statistics when moving from Stade Rennais to Liverpool, based on team and league playing style, (b) We also incorporate the difference in team and league quality to determine the difficulty for the player to transition into the new team (c) Based on our dataset, we give a RAG (red, amber, green) status of our confidence based on the data used for the predictions, and (d) We also enable specific analysis of a selection from our 13 KPI statistics which show how our transferred player (red) compares to other players in their positions across the entire league (grey) and transfer team (orange).*

Our Transfer Portal model is trained to accurately predict thirteen (13) different player-level offensive and defensive outputs, aggregated to per 90 minute (per 90) metrics (shots, Expected Goals (xG), Expected Assists (xA), take-ons, crosses, penalty area entries, total passes, short passes (<32m), long passes(≥32m), passes in attacking third, and defensive actions in own, middle and opposition third). The training set utilizes 26,000 samples of both transfer and non-transfer data across 32 domestic leagues since 2017, where targets are the per 90 metrics of the first 1,000 minutes of a player at a new club, or the next 1,000 minutes if the player remains at their current club. The 1,000-minute limit can easily be changed to predict longer term performance. We evaluated our model against 2,659 historic transfers and 8,677 non-transfers and compared predictions with a baseline model which assumes each player continues to perform as they did before the transfer (i.e., predicted values for each target metric are equal to the most recent player rolling average before the transfer).

We chose to predict the 13 key performance indicators instead of an overall monetary value for two reasons: a) it is quantifiable and reasonably consistent throughout leagues when compared to transfer fees which tend to be quite random and volatile, and b) it is also multi-dimensional which enables decision makers to weight the attribute which is most important to them.

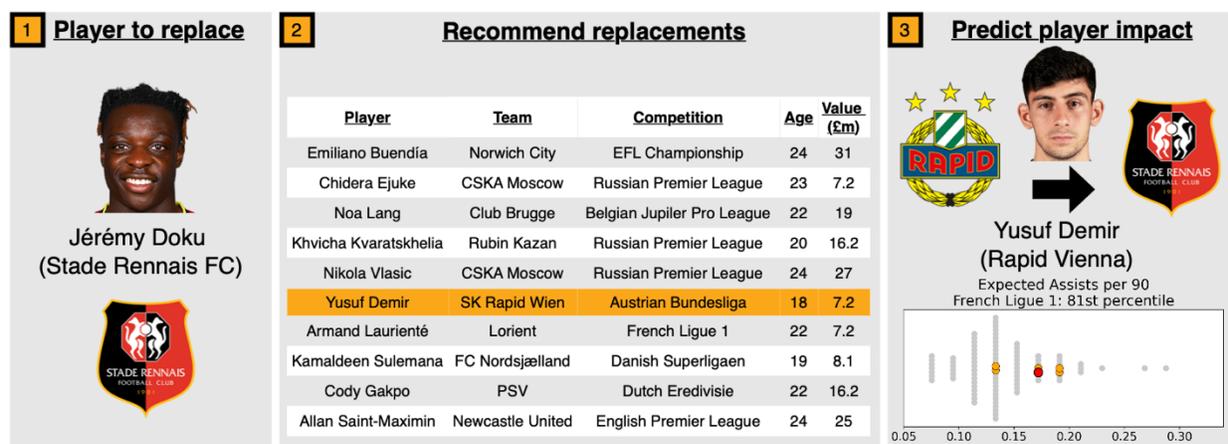

*Figure 2: Our player recommendation system, which requires: (1) A player that is intended to be replaced, (2) Our system then generates a ranked list of top candidates to replace the player, and (3) Specific forecasts of the new player and the impact it will have on the team. In the example shown, we show how Yusuf Demir from the Austrian Bundesliga league would performance at Stade Rennais (date of simulated transfer: 12th July 2021).*

The architecture of our Transfer Portal model is depicted in Figure 3. To build a framework for predicting these player metrics at a new team and/or league, we must represent player, team, and league entities in a personalized feature space which updates after each game played. This is the second module in our approach and in many ways, is the most important part of our prediction pipeline. Without accurate representation of players, teams and league that can update over time, we cannot expect reasonable predictive performance from any modelling approach. Furthermore, we require an approach that can handle low data quantity players and teams, such as breakout youth players or newly promoted teams. To handle these challenges, we craft features that measure both the change in style and ability of the teams and leagues involved in a transfer, in addition to the player's performance relative to other players on their current team. We also utilize a set of "adjustment models" that predict initial feature values for low data players and teams to be used as prior information, which are updated as we collect more data.

In the following section, we go into detail on how these representations are created. In Section 3, we then show how we trained and evaluated our model. In Sections 4 and 5, we show examples of our model in action, from more detailed transfer target analysis in Section 4 to a simpler application of 'Hot or Not' for transfer rumours in Section 5. To our knowledge, no one has been able to execute such an approach in soccer and we highlight relevant works as well as summarize key findings in Section 6.

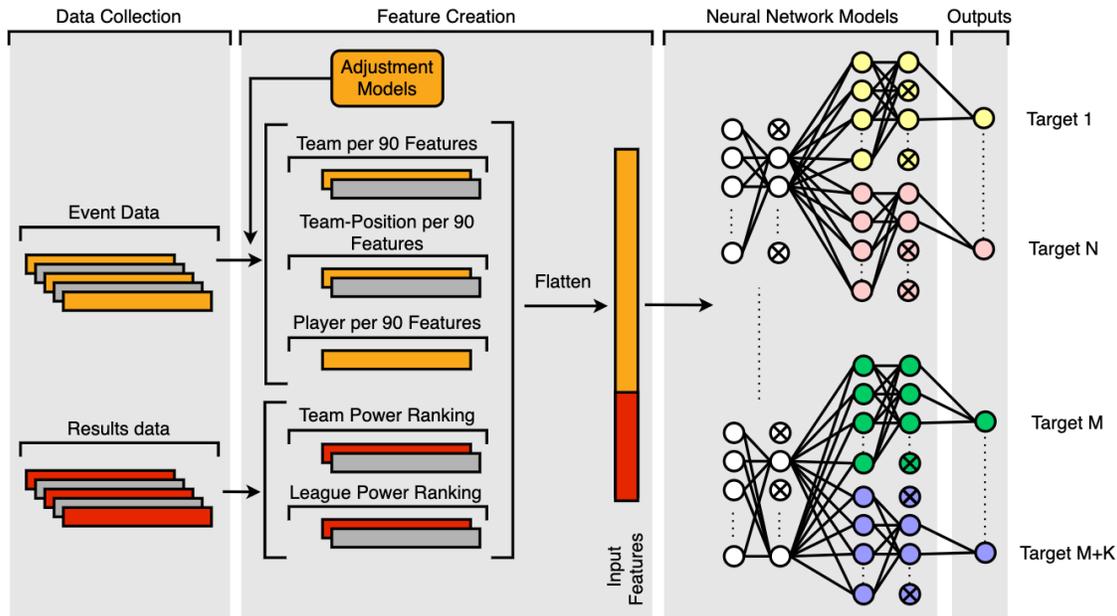

*Figure 3: Our Transfer Portal approach consists of 4 modules: 1) Data collection utilizing spatial event data across all leagues across the globe which captures player performance, as well as result data, 2) Feature creation which captures team, role, player as well as team and league rankings—these features are then adjusted based on personalized predictions when insufficient data is available, 3) Modelling using a deep neural networks, 4) Output feature generation.*

## 2. Player, team, and league feature creation

At a high-level, the Transfer Portal prediction task can be seen as a matrix-completion problem where we aim to predict the "player metrics" features of a specific player to a target team, given the player metrics from the current team, with the additional context of team/league ability, team metrics and team-position metrics for both the current and target team (see Figure 4). Although the intuition of the approach is quite straight forward, to build up the feature representations to achieve this is a major contribution of the paper.

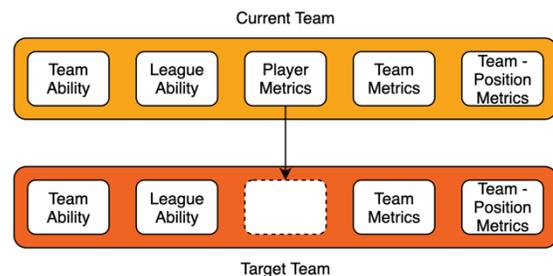

*Figure 4: Our Transfer Portal approach works by predicting the player metrics on the target team, given the player metrics on the current team as well as the team, league ability and team metrics (including per position).*

As depicted in Figure 4, the feature modules we created were: team and league ability features, normalized player metrics, normalized team metrics and position-based team metrics. In this section we go into detail on how we created these features. Before those features could be created however, we first had to utilize the raw event based-data which contained spatial and temporal information of the on-ball events which we utilized from Opta [2].

### 2.1. Raw player and team features

The first stage of our feature pipeline is to aggregate event-level to game-level data. Our Transfer Portal model requires features and targets on a per 90 scale, for data at a player, team, and team-position level. We do this by counting events at the lowest level, by player-position. We use formation event information to update the position of each player and count events and time played in each position. Take for example, Jérémy Doku playing 85 minutes as a winger and 10 minutes as a central midfielder for Rennes in a single game. We ensure that events from Doku playing as a central midfielder are counted separately to those as a winger.

From there it is an aggregation exercise to find the total minutes played and total action counts for each position in the team. For example, counting the minutes and shots from all central midfielders that played for Rennes in that match, including Doku's data. Finally, the team level metrics aggregates the position level information, so combines the minutes and actions of all positions for Rennes in that match.

## 2.2. Team and league ability weightings

One of the most important inputs for accurately predicting player performance is a measure of ability for both the teams and leagues involved in a transfer. Clearly, one would expect players moving to lower quality leagues to see a bump in their performance as they play against lower quality opponents. However, this bump might also need to be factored against the change of team ability relative to the league, since although the player might be moving to a lower quality league, they might also be moving from a top of the table team to a relegation candidate.

To do this, we need to create an "ability feature" which provides an accurate measure of team quality. But just as importantly, the measure must but be flexible enough to adapt quickly to the changing ability of teams over time and cover teams from as many leagues as possible. This will provide us with the resources to retrain and grow the Transfer Portal model across more leagues over time. To enable a truly global rating system which reacts to the changing football landscape, we introduce a hierarchical approach, which enables propagation of ability scores across all continents, countries, and leagues. This approach has enabled us to record daily ability ratings since 1990 across 195 countries, 423 leagues and over 20,000 teams. As far as we are aware, this is the largest soccer ratings system that exists. Our hierarchical approach is shown in Figure 5.

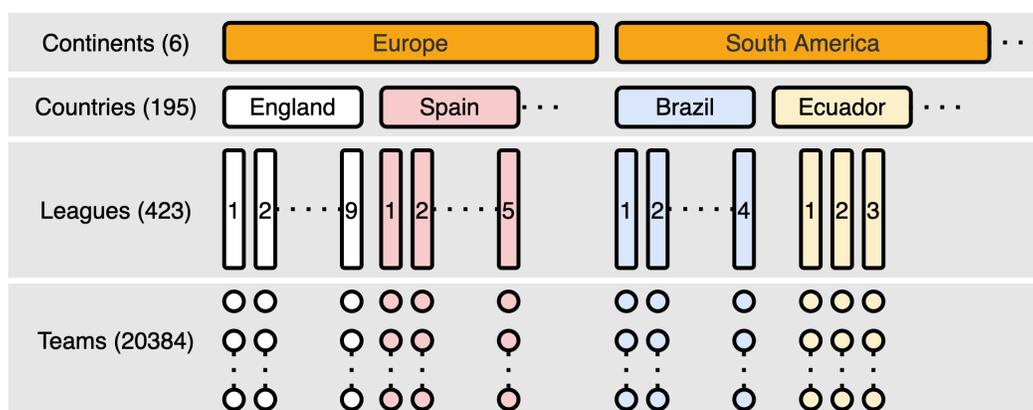

*Figure 5: Our four-level hierarchy approach assigns ability ratings to each continent, country, league, and team).*

These ability scores are based on the Elo rankings originally used in chess [3]. Elo ratings provide a simple approach for updating team ability ratings after each game. The expected result of each match, which is based on the pre-game Elo difference between the two teams, is compared to the actual result of the match. Based on the difference in expected and actual results, both teams will have their Elo rating adjusted. For example, if the home team is expected to lose, but wins, then they will see an increase in their Elo score.

Given York City FC in the 6th Tier of England (National League North) as of 2021, their final ability score is a sum of 4 separate Elo ratings across their continent, country, league, and within league team values as shown below:

$$E_{York\ Final\ Score} = E_{York\ within\ league} + E_{National\ League\ North} + E_{England} + E_{Europe}$$

The base team level of the hierarchy has its Elo adjusted for every game. However, the league, country and continent levels are only adjusted when a game takes place between teams in different groups of that hierarchy. We only ever adjust the highest level of the hierarchy affected.

To understand the value of our approach, it is best illustrated via the following example. Take the 2019 Club World Cup final between Liverpool FC (Premier League) and Flamengo (Campeonato Brasileiro Série A). Based on the result, we update both the Liverpool and Flamengo within league team Elo ratings, along with the European and South American Elo ratings. We do not touch the league or country Elo's as we only touch the highest level of the hierarchy affected (apart from within league team score which is always updated). This game would therefore impact all teams in Europe and South America, as their final ability score is a sum of the hierarchies they fall into, which include the European and South American groups which have their abilities scores adjusted based on the results of this game.

The final part of our ability score process is to scale the team ability scores between 0-100. We scale all teams daily to prevent the impact of potential Elo inflation in the system over time, and means that on any given day, the best team in the world will have score 100, and the worst will have score 0. This provides us with a single value which enables team comparison for any target transfer globally, which we denote our Power Ranking scores.

Take for example one of the biggest 2021 transfers, FC Barcelona's purchase of Sergio Agüero from Manchester City in May 2021. In Figure 6, we show our Power Ranking ability ratings of both teams for the 10 years preceding the transfer. We also include Agüero's original team, Club Atlético Independiente of the Argentine Primera División, to show how these values are spread globally and can be used to inform transfer ability change across any league.

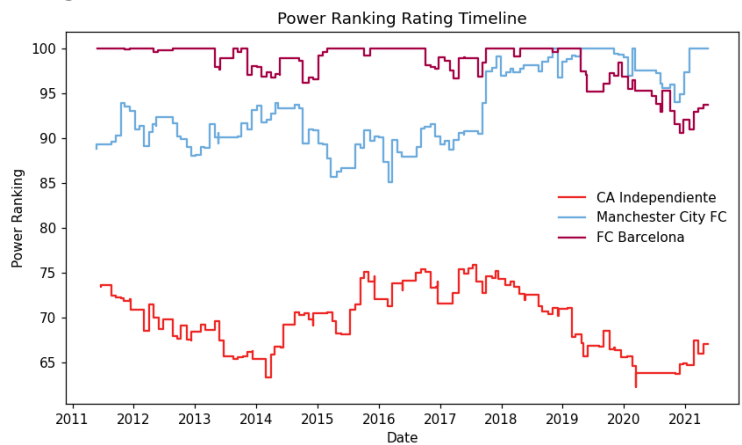

*Figure 6: We show our power rankings for Independiente, Man City and Barcelona over the last 10 years.*

According to our rating system, between 2011 and 2019 FC Barcelona had almost always been a higher ability team compared to Manchester City. However, at the time of the transfer in May 2021, FC Barcelona were at their biggest deficit to Manchester City over the previous 10 years. This is extremely important information if our Transfer Portal model is to learn how transfer performance is impacted by team ability. However, without league context this information can be misleading, so we also need to understand the ability of other teams within the leagues we are targeting. In Figure 7, we show the distribution of team Power Rankings within the English Premier League and Spanish La Liga at the time of the Agüero transfer in May 2021

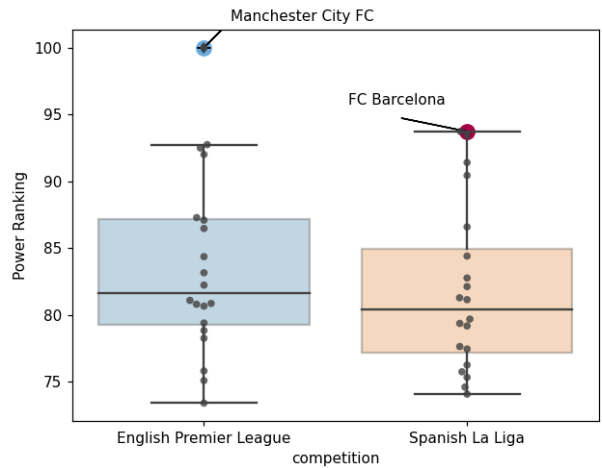

Figure 7: Comparing the power ranking distributions between the EPL and La Liga in May 2021 at the time of Sergio Aguero's transfer.

The two leagues are very evenly matched; however, the English Premier League appears to slightly stronger league overall. Having said this, whilst the transfer is to a very slightly lower ability league, FC Barcelona are less of an outlier in terms of ability compared to their league. Again, this is very important information that the model must use to train on previous transfer performance and predict how future transfer will perform. Moving to a slightly lower quality league might help a player, but we also need to understand that FC Barcelona might not have the quality advantage over opponents that Agüero has recently enjoyed at Manchester City.

### 2.3. Normalized per 90 minute team and player features

Following the representation of team and league quality, we now require more detailed descriptions of the teams and individuals involved in the transfer. As with the prediction targets, we do this using per 90 features that describe both the style and performance of teams and players. At a player level, we use the 13 metrics which are aggregated across their previous games. To enable updating features after each game, we utilize rolling window averages over a set number of minutes played. For example, to represent the number of passes a player makes per 90 minutes of play, we use the number of passes they have made in their previous N minutes, where N is a model constant chosen to be 1000 in this paper which ensures we are always utilizing a time window of the most up-to-date information and discarding old data. We also aggregate the 13 metrics across historic team level data, to create team data, and across different positions to create team-positional data.

For example, when predicting how Sergio Agüero would perform at FC Barcelona, we would want information on how both FC Barcelona and Manchester City use their strikers. To do this, we would aggregate all striker data from the past $M$ minutes (we take $M = 3000$ in this paper) at both teams, to create the 13 team-position level features for this transfer.

### 2.4. Adjustment models

The key to our player and team feature approach is how to represent low-data entities. It would be easy to overlook or exclude transfers where we do not have substantial game time for either the player or team in question. However, it is often the case that these transfers are the most important. Take for example, Emile Smith Rowe at Arsenal. He broke onto the scene in the second half of the 2020 season to become one of the brightest young players in the English Premier League. As a prospective buyer, we would want to predict how Smith Rowe would perform at our club, even though he has only played a handful of games.

Of course, this will never be possible to the degree of confidence that we would expect from a prediction for a well-established player, but we still want to make an educated decision. This brings us back to our rolling window averaging that we are using for the team and player features. Once we have substantial game time, this approach works well. However, in many cases we do not have the luxury of such data, which can cause problems in several cases. The Smith Rowe example with a youth player who has only played in a couple of matches. How do we represent this player since their raw data will be extremely noisy? A similar problem would be a player who has recently moved to a new club where they only have a handful of minutes played. We also have this problem at a team level. Potentially the team has just been promoted from a league with no data, or they rarely play the position for which we want to simulate a player in. Or more commonly, the team has just moved leagues and we are yet to see how they will perform against their new standard of opposition.

To solve these problems, we utilize internal models to predict initial performance of data poor players and teams to be used as prior information, from which we update the features as we collect more data. We use the term prior in a loose sense, it is simply an initial value for the features which update as we collect more data to obtain a more accurate understanding of the entity's performance. To calculate the final feature value, we use a linear weighted average between the prior and raw rolling window average. Initially the weighting will be heavily in the prior's favor, but over time as we collect more data the weighting moves towards and then completely onto the rolling window average.

Denoting feature $i$ for player-position-team-league $j$ at game $g$ as $X_{i,j,g}$, we can explicitly define this as
$$X_{i,j,g} = (1 - w_{j,g})P_{i,j} + w_{j,g}R_{i,j,g}$$
where our weighting $w_{j,g} = \min(1, \sum_{t=0}^{g} m_{j,t}/c)$ is the minimum of 1 and the sum of minutes played $m$ by the player-position-team-league $j$ in all their games up to game $g$, divided by some user defined constant $c$. Finally, $P_{i,j}$ is the prior value for player-position-team-league $j$ in feature $i$, and $R_{i,j,g}$ is the raw rolling window average of feature $i$ for player-position-league-season $j$ at game $g$. By controlling the constant $c$, we can adjust the speed at which the weighting shifts from the prior to the rolling average.

These feature adjustment models require that our features are calculated in order of: i) Ability score, ii) Team/Team-Position, and iii) Player. To provide a nested approach to calculating the priors of features in each category. For example, team feature prior predictions require team and ability score data, whilst player feature prior predictions require player, team, and ability score data. We save the details of the adjustment models for the Appendix.

A player for which these adjustment models are necessary due to them moving both teams and leagues recently is Ismaïla Sarr. Figure 8 shows an example of the Expected Assists (xA) per 90 feature values for Ismaïla Sarr playing as a winger, with $c = 1000$. In each plot, we follow his team and league progression, from Rennes in Ligue 1 to Watford in the English Premiership and then English Championship. Each time the team or league changes, we recalculate the prior value (yellow) of the feature using our player adjustment model which will consider historical information of the player, team, and league. The blue dotted line shows the raw rolling xA per 90 values up to the past 1000 minutes, and green is the feature value for Sarr. We can see how the weighting shifts towards the raw rolling xA per 90 as Sarr gets more playing time as a winger in his current team and league combination.

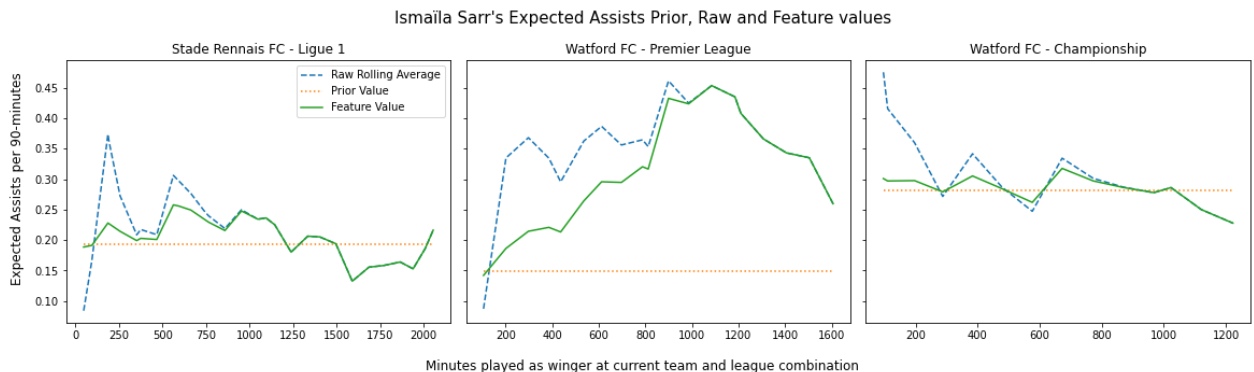

*Figure 8: Prior, raw data and actual feature values for Expected Assists per 90 minutes for Ismaïla Sarr.*

Our player adjustment model underestimates Sarr's xA per 90 at Watford in the Premier League, however this data provides the player adjustment model with newfound knowledge which is then used to inform the prediction of Sarr in the English Championship. This prior value then accurately calculates a high-performance level which is matched as we collect more information throughout his time in this league. To ensure that our confidence in feature values is conveyed to the user, we use a simple RAG (red, amber, green) system to denote whether the features are highly dependent (red), slightly dependent (amber) or not at all dependent (green) on our predicted prior value.

## 3. Transfer Portal prediction model

### 3.1. Model architecture

Having described the feature pipeline of our Transfer Portal in the previous section, in this section we detail the prediction model which takes these input features and translates them into predictions over 13 unique per 90 target metrics. For modelling we utilized a grouped feature structure where related targets—for example, xG and shots per 90—were modelled together using a multi-head approach. This enabled us to use unique subsets of input features that were relevant to the targets in each group, to share information across the prediction targets, without overloading the model with less relevant data that in many cases introduced noise and negatively impacted on predictive model performance. Across our 13 targets, we fit 4 separate models, with Table 1 showing the groupings.

*Table 1: Grouping structure of target variables.*

| Group Number | Targets |
|---|---|
| 1 (Shooting) | Shots, Expected Goals (xG) |
| 2 (Passing) | Expected Assists (xA), Crosses, Total Passes, Total Short Passes (< 32m), Total Long Passes (>= 32m), Passes in Attacking Thirds, Penalty Area Entries |
| 3 (Dribbling) | Take-ons |
| 4 (Defending) | Defensive Actions in Own Third, Defensive Actions in Middle Third, Defensive Actions in Opposition Thirds |

For each target group we fit a multi-head neural network model using Tensorflow [4]. In each case, we utilise a dense initial layer of all features for the target group, before splitting into individual layers for each target as shown in Figure 9.

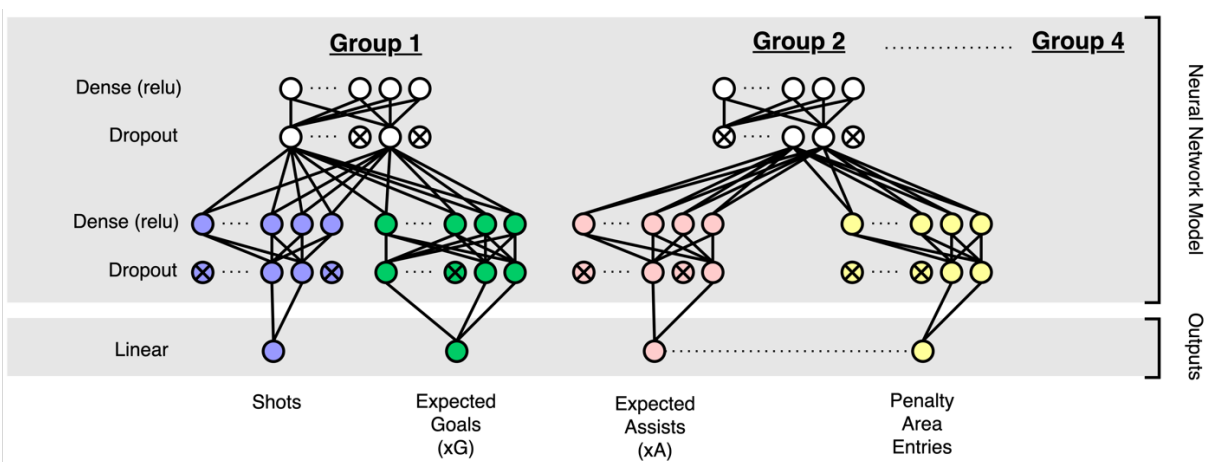

*Figure 9: Representation of the four grouped Neural Network models used by the Transfer Portal.*

This structure enables the sharing of relevant predictive information using out initial dense layer, before splitting out into uniquely optimised layers for each target. We optimised several hyperparameters (learning rate; batch size; dropout and number of neurons in each hidden layer) over a large search space using the Bayesian hyperparameter optimisation library, HyperOpt [5].

## 3.2. Model performance

The Transfer Portal model was trained over 26,000 samples of both transfer and non-transfer data, where targets were the per 90 metrics of the first 1000 minutes of a player at a new club or the next 1000 minutes if the player remained at their current club. The 1000-minute limit can be changed to predict longer term performance.

We evaluated our model against 2659 historic transfers and 8677 non-transfers and compared predictions with a baseline model which assumes each player continues to perform as they did before the transfer (i.e., predicted values for each target metric are equal to the most recent player rolling average before the transfer).

On average across the 13 metrics, we see a 49% improvement in mean squared error versus the baseline model when only considering transferred players. This is reduced to a 21% improvement when including both transfer and non-transfer test data, which is to be expected as historic player performance is a better predictor when staying at the same team. In Figure 10, we show the prediction for player xG per 90 across the baseline and Transfer Portal model for the 2659 transfers. We can see via the calibration plot on the right-hand side that there is slight over prediction for the lowest xG players and a slight under prediction for the highest xG players. However, there is a large 54% reduction in the mean squared error of these predictions, showing the huge value gained from including team style and ability information to inform future player performance.

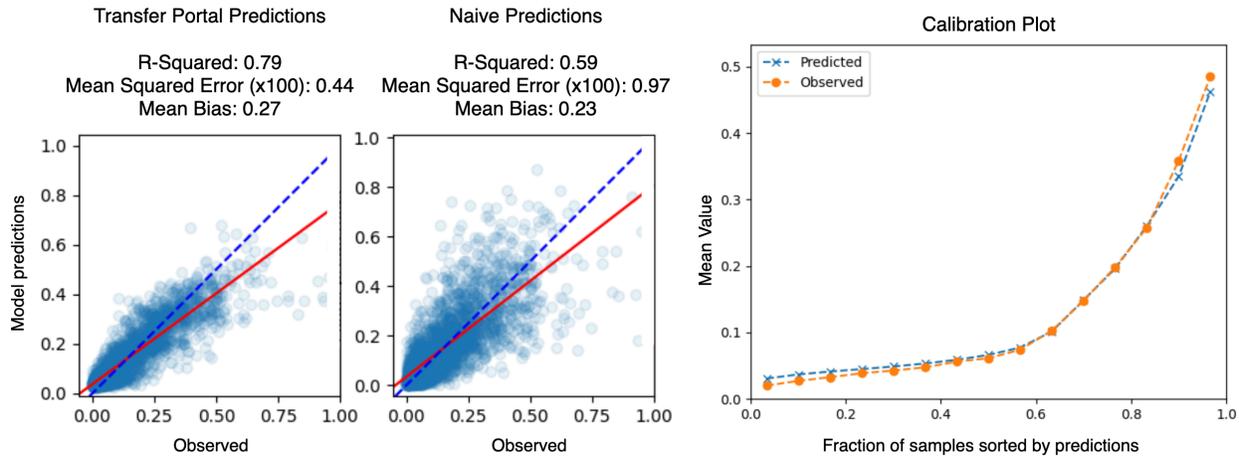

*Figure 10: Left: Comparison of Transfer Portal and baseline (current player performance) predictions for Expected Goals (xG) per 90 minutes for transfers only. Dotted blue line is perfect predictions and red line is line of best fit through the predictions. Right: Calibration plot for xG predictions.*

Some metrics show a greater improvement than others, as they are influenced more by team ability and style. The percentage improvement of mean squared error against the baseline model for each of the 13 targets across all transfer test data is displayed in Table 2, with the improvements ranging from 37% for crosses per 90 to 61% for short passes per 90.

## 4. Model outputs

We have now detailed the structure of our Transfer Portal model and shown how it improves our prediction of future performances for transferred players. In this section we show several possible applications of our Transfer Portal applied to a real-life example of Stade Rennais FC recruiting a right-sided winger during the 2021 summer transfer window—a possible replacement for their first-choice right winger Jérémy Doku who was linked with a move to several big clubs following a successful 2020 UEFA European Championship. Firstly, we generate a shortlist of potential recruits using a customized score which allows the user to weight the importance of specific metrics (e.g., Expected Assists (xA) and take-ons more important, defensive actions less important), in additional to other simple filters (e.g., age, value, league, etc.). Secondly, we examine more closely the simulated performance of three shortlisted players: Kamaldeen Sulemana, Yusuf Demir, and Noa Lang. Finally, we switch perspective to look at Jérémy Doku's possible outbound transfer and simulate his performances across various linked teams.

*Table 2: Comparison of our Transfer Portal predictions vs baseline.*

| Target (per 90-minutes) | Transfer Portal Mean Squared Error | % Improvement over baseline |
|---|---|---|
| Shots | 0.16 | 46% |
| Expected Goals (xG) | 0.0044 | 54% |
| Expected Assists (xA) | 0.0013 | 54% |
| Crosses | 0.84 | 37% |
| Total Passes | 30.09 | 57% |
| Total Short Passes (<32m) | 25.24 | 61% |
| Total Long Passes (>=32m) | 1.29 | 40% |
| Total Passes in Attacking Third | 4.89 | 48% |
| Penalty Area Entry Passes | 2.95 | 41% |
| Take-ons | 0.40 | 40% |
| Defensive Actions in Own Third | 0.79 | 46% |
| Defensive Actions in Middle Third | 0.37 | 49% |
| Defensive Actions in Attacking Third | 0.035 | 49% |

## 4.1. Generating a shortlist

Our Transfer Portal simulates the performance of a transferred player across a total of 13 metrics; although we could simply order a list of players by a single predicted metric (e.g., highest xG per 90), we may wish to evaluate prospective transfers more holistically across a range of metrics. We therefore create an overall score based on a set of custom weightings which allow the user to quantify the importance of each metric; for example, for an attack-minded winger, we may be more interested in goals and assists than in defensive actions.

*Table 3: Custom weightings used to generate a shortlist of right wingers ordered by an overall similarity score, roughly based on the performance profile of Doku at Stade Rennais FC.*

| Target | Weighting |
|---|---|
| Take-ons | 1.0 |
| Expected assists (xA) | 1.0 |
| Expected goals (xG) | 0.7 |
| Crosses | 0.2 |
| Penalty area entry passes | 0.2 |

In brief, each predicted target is mean normalized and multiplied by a user-defined weighting between 0-1, with a final score between 0-1 derived by summing weighted scores and dividing by the sum of the weights. Based on a set of weightings which roughly correspond to Jérémy Doku's main strengths at Rennes (Table 3), we generate a shortlist of 10 players (Table 4) which additionally satisfies the following criteria:

- Players under 25 years old (on 12/07/2021)
- Estimated transfer value of less than £30 million, via Transfermarkt [6]
- Minimum of 450 minutes played as a winger in the last 365 days
- Capable of playing at right wing even if left wing is primary position, via Transfermarkt [6]
- Teams with less than 2400 Power Ranking and in one of 16 top European leagues

*Table 4: Shortlist of 10 wingers most suitable for Stade Rennais. Score is a weighted average of several per 90 minute metrics using custom importance sliders (Table 3). Filters are also applied, such as age, estimated value, and team Power Ranking.*

| Similarity Score | Player | Team | Competition | Age | Value (£m) |
|---|---|---|---|---|---|
| 0.625 | Emiliano Buendía | Norwich City | EFL Championship | 24 | 31 |
| 0.622 | Chidera Ejuke | CSKA Moscow | Russian Premier League | 23 | 7.2 |
| 0.614 | Noa Lang | Club Brugge | Belgian Jupiler Pro League | 22 | 19 |
| 0.597 | Khvicha Kvaratskhelia | Rubin Kazan | Russian Premier League | 20 | 16.2 |
| 0.597 | Nikola Vlasic | CSKA Moscow | Russian Premier League | 24 | 27 |
| 0.592 | Yusuf Demir | SK Rapid Wien | Austrian Bundesliga | 18 | 7.2 |
| 0.591 | Armand Laurienté | Lorient | French Ligue 1 | 22 | 7.2 |
| 0.589 | Kamaldeen Sulemana | FC Nordsjælland | Danish Superligaen | 19 | 8.1 |
| 0.585 | Cody Gakpo | PSV | Dutch Eredivisie | 22 | 16.2 |
| 0.583 | Allan Saint-Maximin | Newcastle United | English Premier League | 24 | 25 |

## 4.2. Examining candidates

In a real recruitment environment, an automated shortlist would be supplemented by traditional scouting methods to identify the most optimal candidates using a range of more thorough but less quantitative criteria such as live assessment and video evaluation. We imagine a similar process applied for the purposes of our example and hone our shortlist down to three players whose predicted performance we will more closely examine: Kamaldeen Sulemana, Yusuf Demir, and Noa Lang.

Our approach to comparing predicted performance for a player at a new team against their current club, is to simulate them at both their current and target team for the next 1000 minutes played, where, as we discussed in Section 3.2, this minute value can be changed depending on the application. To provide a baseline with which to compare simulated transfer performance, we generate performance predictions using Transfer Portal for players at their current club too—as opposed to using their actual observed performance measures at their current club. We prefer both sets of predictions to be generated by the same underlying process and find the predictions of our model less sensitive to noise than observed data.

To visualize the comparison between player performance at their current and transferred team, we use swarm plots to add context to our predictions. Figure 11 shows how we can do this with a simple example looking at only predicted shots per 90 for Cody Gakpo of PSV Eindhoven in the Dutch Eredivisie moving to Stade Rennais as a winger. Gakpo is simulated at his current club in the left plot, and highlighted in red. His PSV winger teammates are highlighted in orange. Finally, we simulate all other wingers in the Dutch Eredivisie at their current clubs in grey. Therefore, we now have context on how Gakpo compares to his current team and league-mates. We do the same in the right plot except for the destination team and league, which in this case is Rennes. Hence, we are comparing how Gakpo is expected to perform in comparison to his new winger teammates and league-mates in France. In both cases, to help with interpretation, we include the league percentile value for league ability context. In this example, Gakpo was slightly above league average for shots per 90 at PSV— the 59th percentile of wingers in the Dutch Eredivisie—which is significantly reduced to the 27th percentile of wingers in Ligue 1 when moving to Rennes.

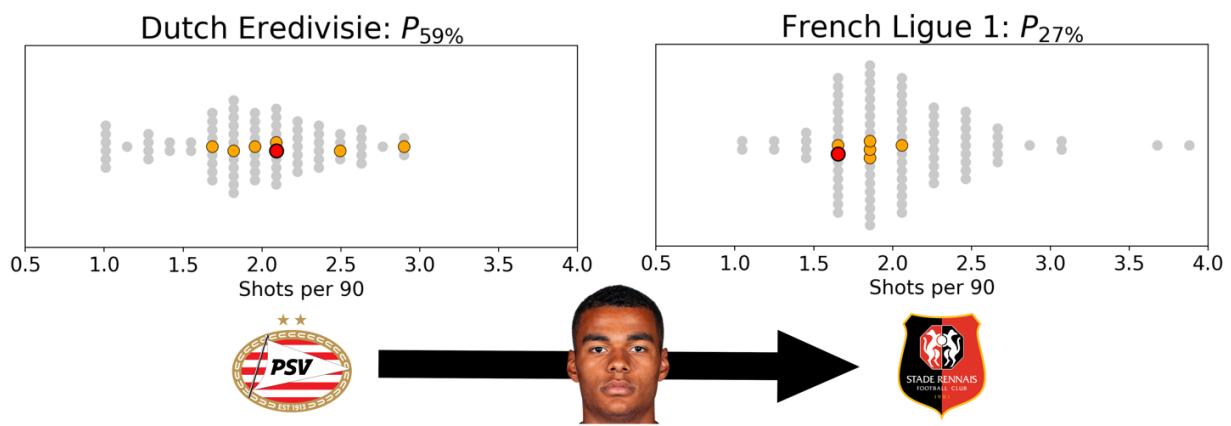

*Figure 11: Example of how we use swarm plots to visualize predicted transfer performance with team and league context, using Cody Gakpo simulated to Stade Rennais FC from PSV Eindhoven.*

### 4.2.1. Kamaldeen Sulemana (FC Nordsjælland)

In Figure 12 we show our prediction dashboard for Kamaldeen Sulemana. He appears to be a very attack-minded player, with Figure 12d showing his predicted performance at current club FC Nordsjælland being in the top 12% of shooting metrics and top 1% of take-ons compared to other wingers in the Danish Superliga, but the bottom 20% of passing metrics including crosses and xA. This is partly explained by FC Nordsjælland's playing style: other wingers at the club (orange, Figure 12d) have very high numbers in shooting metrics and low numbers in passing metrics compared to the rest of the league. Sulemana's elite shooting outputs are predicted to be attenuated somewhat at Rennes (e.g., over 20% reduction in xG, Figure 12a), which is partly explained by it being a transfer to a more difficult league (Figure 12b), and the fact that Rennes wingers have relatively average shooting outputs compared to the rest of Ligue 1 (Figure 12d). By the same token, some of Sulemana's passing metrics are expected to be boosted somewhat in Ligue 1—despite the greater difficulty of the league—as Rennes wingers have a higher average number of xA, crosses, and penalty entries. Finally, Sulemana's huge number of take-ons (>50% more than anyone else in the Danish Superliga) are mostly retained at Rennes, although are less of an outlier in Ligue 1.

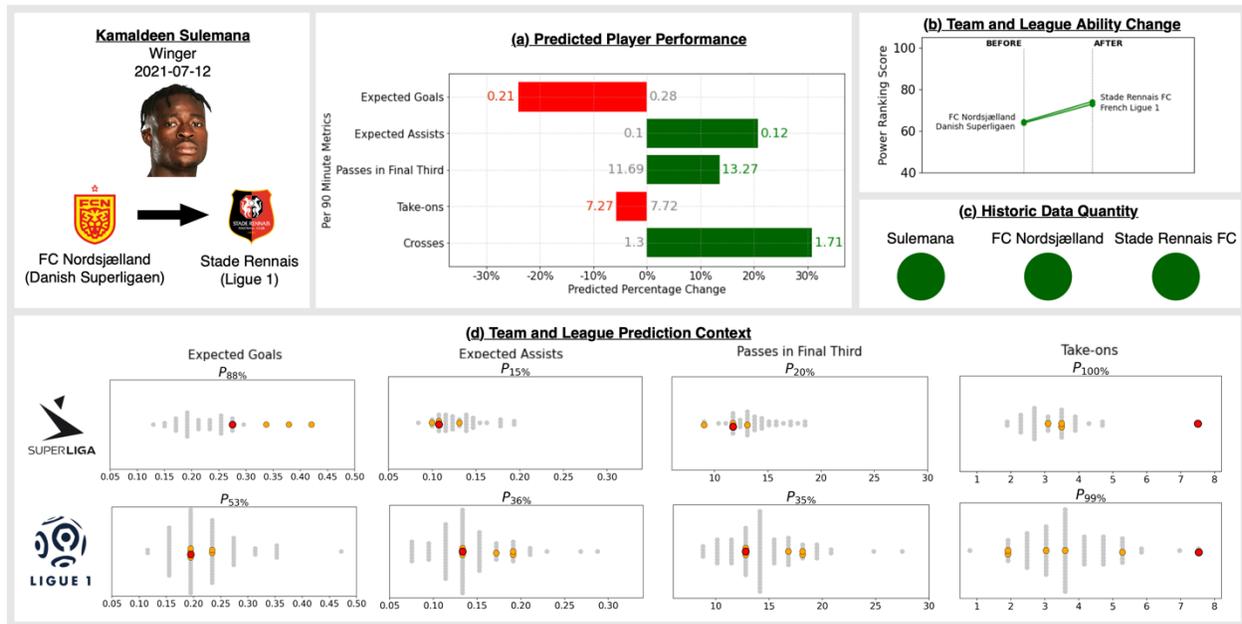

*Figure 12: Example dashboard of Kamaldeen Sulemana's simulated performance as a winger at Rennes.*

### 4.2.2. Yusuf Demir (Rapid Vienna)

Figure 13 shows that Yusuf Demir at Rapid Vienna has a more rounded profile than Sulemana, being in the top 26% of wingers in the Austrian Bundesliga in all predicted shooting, dribbling, and passing metrics (Figure 13d). Most of his performance metrics are attenuated by a proposed move to Rennes as a result of the more challenging league (Figure 13b)—for example, his xG per 90 is predicted to be reduced to that of a median winger in Ligue 1. However, in terms of the two metrics we are most interested in according to our customised rating—xA and take-ons—Demir is still predicted to rank in the top 20% of wingers in Ligue 1 and one of the top 2-3 at Rennes (orange, Figure 13d). It is also important to note that we highlight Yusuf Demir with a red light (Figure 13c) to highlight the fact that the then 18-year-old has less than 500 minutes played as a winger as of the transfer date; this is to emphasise that there is a greater degree of uncertainty in his features than a player with more minutes.

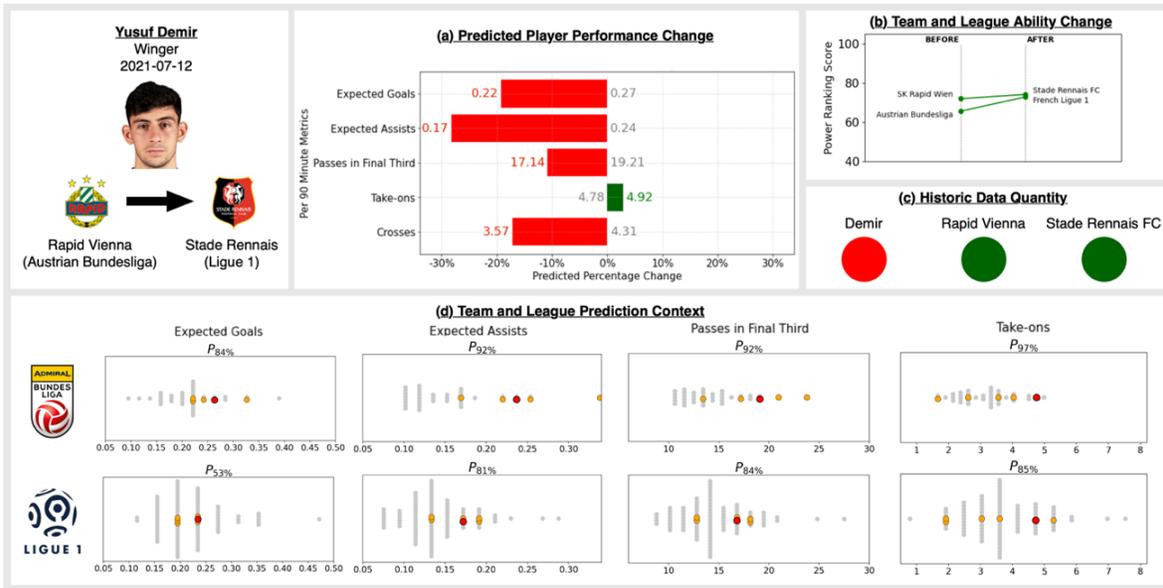

*Figure 13: Example dashboard of Yusuf Demir's simulated performance as a winger at Rennes.*

### 4.2.3. Noa Lang (Club Brugge)

In   Figure 14d, we see that Noa Lang at Club Brugge also excels in each of passing, dribbling, and shooting—although has a lower engagement in crossing than other wingers. A transfer to Rennes from the less competitive Belgian Jupiler Pro League (where Jérémy Doku arrived from) is met with a reduction in several predicted metrics—his xG, for example, is reduced from being the very highest of wingers the Jupiler Pro League to the 88th percentile in Ligue 1. However, his elite numbers in terms of xG, xA, and passing at Club Brugge are predicted to be retained despite the relative difficulty of Ligue 1—partially because of Rennes' playing style—with each of these predicted to be in the top decile of wingers in Ligue 1. An important note is that Lang is right footed and tends to play on the left wing as an inverted winger—cutting inside to shoot or pass rather than crossing from the wing; as such, he would not provide a good fit if we were trying to replace a traditional winger on the right.

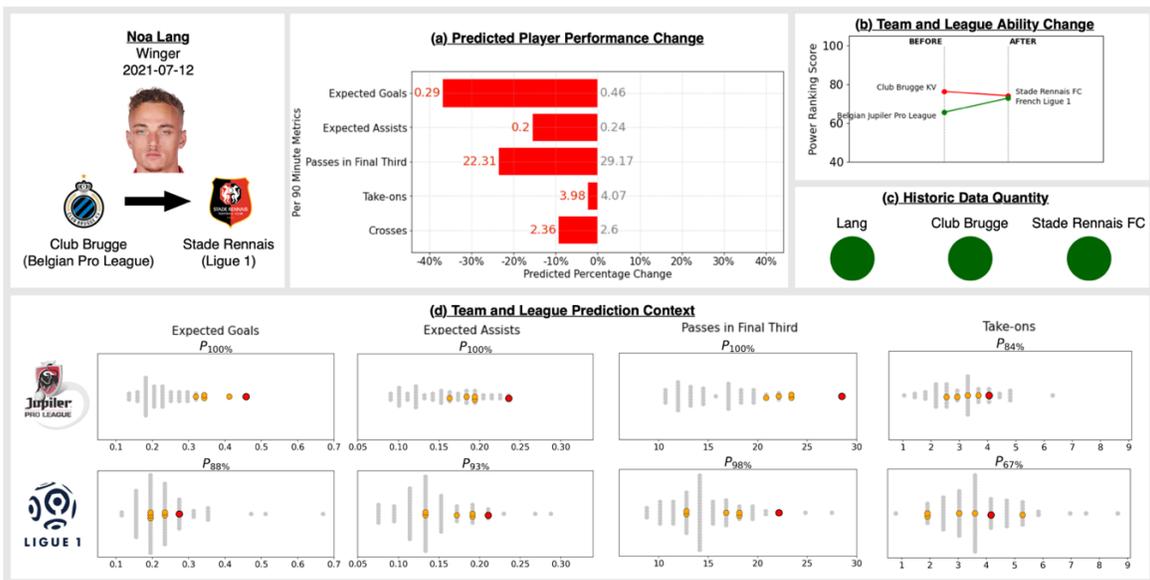

*Figure 14: Example dashboard of Noa Lang's simulated performance as a winger at Rennes.*

## 4.3. Predicting Doku's next move

Having shown how Rennes could look to replace their star player Jérémy Doku, we can now switch the application of our model to explore how Doku might performance at clubs who have shown interest in signing him. From the perspective of the selling club, this application could help convince the departing player to choose the club which offers the best chance of success or help loan managers choose the best destination for an emerging talent to develop whilst out on loan.

The flexibility of the Transfer Portal model means that at present, we can predict players across 32 domestic leagues, a value which will be rapidly increased as the model is validated across more leagues. Therefore, the model can be used to seek players from better value leagues, which might provide lower-budget clubs with potential bargains—such as Rennes recruiting Doku from Anderlecht. We can also do the inverse and show how players perform switching from higher quality to lower quality teams—the types of moves which are common later in a player's career.

As of December 2021, the frontrunners for Doku appear to be Liverpool FC in the English Premier League and FC Barcelona in the Spanish La Liga. In this section we compare the projected performance of Doku at these two prospective buyers and demonstrate the flexibility of our model with a more unrealistic destination: relegation candidates in the Korean K-League 1, Gwangju FC.

### 4.3.1. Liverpool, Barcelona, or even Gwangju?

In terms of team and league quality, the English Premier League and Spanish La Liga are both higher quality leagues compared to Ligue 1 (Figure 15a) which may be expected to hinder Doku's performance; however, Liverpool and Barcelona are both title-contenders in their leagues which should greatly benefit Doku's attacking metrics. Gwangju present the opposite scenario, being in a much lower quality league but being relatively poor performers as relegation candidates.

Doku's xG per 90 is predicted to increase significantly at both Liverpool and Gwangju compared to Rennes (Figure 15b). Whilst Barcelona wingers have elite expected goal outputs compared to the rest of the league, Doku's projected numbers remain roughly the same as at Rennes; however, he still ranks at the 69th percentile of wingers in La Liga compared to the 40th percentile of wingers in Ligue 1 because of the stylistic differences between these leagues. The reverse pattern is observed in xA, where Doku's production is predicted to drop slightly at Liverpool and increase at Barcelona (Figure 15c).

The influence of team style upon passing metrics appears to interact with team rating. Doku's final third passes per 90 increases at Barcelona and decreases slightly at Liverpool to roughly the averages of their respective current wingers (Figure 15d). However, although Doku's individual numbers drop slightly at Gwangju compared to Rennes, he is still predicted to pass much more than the other wingers on his team—in the 73rd percentile of K-League 1 wingers. This may be a more realistic representation of the impact of an elite player on a less skilled team: individual production may drop somewhat but is still expected to remain relatively high compared to the rest of the team.

Finally, take-ons are an example of a more individual, as opposed to team-oriented, metric. Although managers can instruct players to engage in more or fewer dribbles, it is a somewhat irreducible element of player style; this can be seen in Figure 15e where the distribution of take-ons within a team varies widely from player to player. As such, Doku's elite number of take-ons per 90 at Rennes is retained across each of Liverpool, Barcelona, and Gwangju with him remaining near the top 10% of wingers.

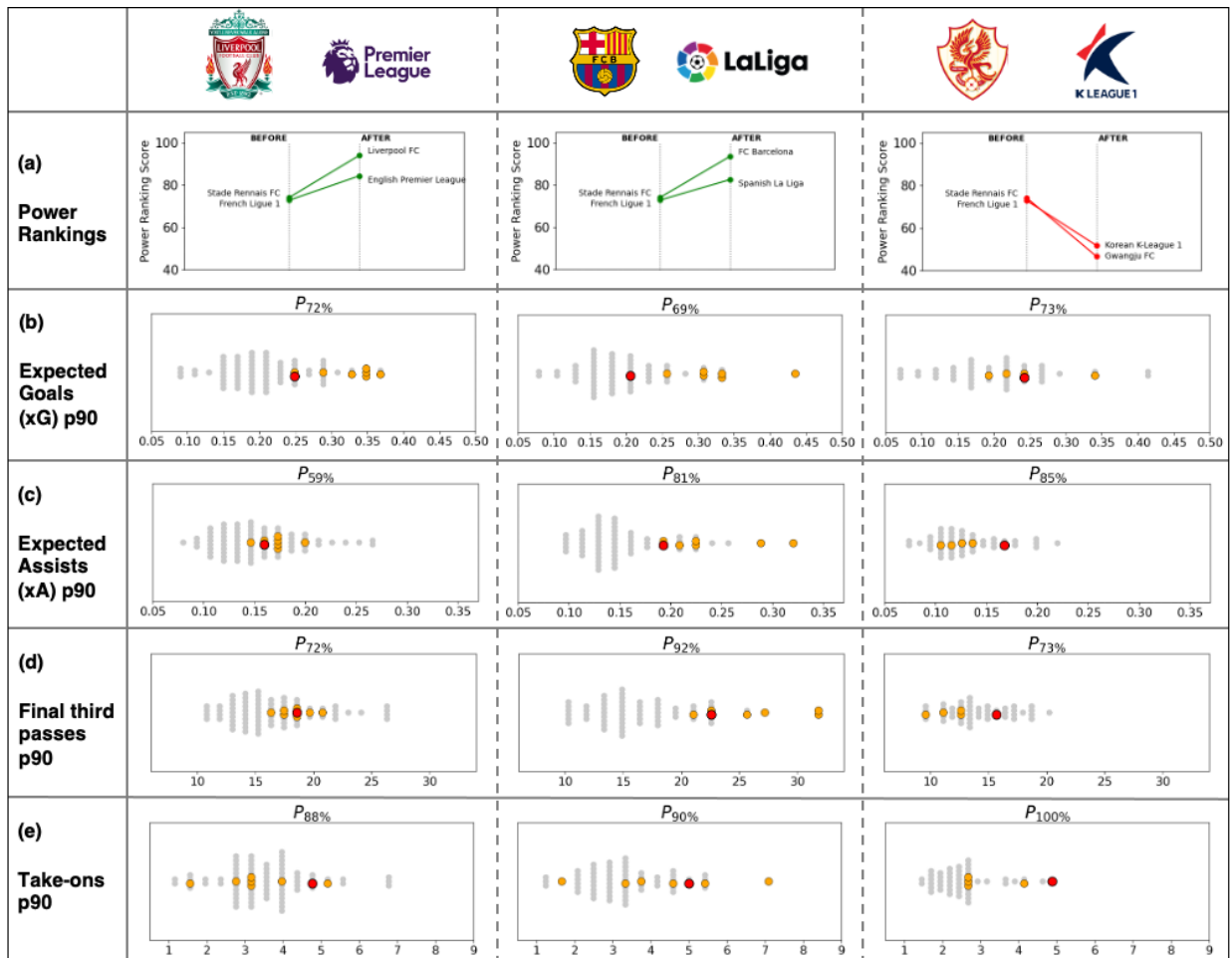

Figure 15: Comparison of Jérémy Doku's predicted performance as a winger at Rennes and three possible destinations: Liverpool (left column), Barcelona (centre column), and Gwangju (right column).

# 5. Transfer Portal for transfer rumours

Another application of our Transfer Portal model is to quickly validate the projected quality of any latest transfer rumour with no information other than a player's proposed destination and playing position. We present here a 'Hot or Not' application using the latest transfer rumours across a range of positions and leagues to suggest whether a player would perform better or worse at an individual level following a proposed transfer—and more importantly whether they would represent a successful purchase for the club.

- **Kylian Mbappé to Real Madrid (Striker): Hot**
  Mbappé's elite performance in Ligue 1—in the 97th to 99th percentile of strikers in most attacking metrics—is expected to continue with Real Madrid in La Liga despite the greater difficulty of the league and an 8-15% decrease in predicted outputs. He is expected to rank at the 94th percentile of strikers in La Liga for xG and xA and the 98th percentile for take-ons.
- **Harry Kane to Manchester City (Striker): Hot**
  Despite a slight decrease in xG out from the 99th percentile to 95th percentile of Premier League strikers, Kane is still predicted to rank as Manchester City's top striker. His xA and take-ons, whilst not a huge part of his game, are also projected to remain roughly the same at Manchester City, suggesting he could be the replacement they need for Sergio Agüero.
- **Frenkie de Jong to Manchester United (Central Midfielder): Not**
  de Jong has an elite level of versatility at Barcelona, ranking in the top 10% of central midfielders in La Liga in terms of xA, final third passes, and take-ons and even in the top 20% in terms of xG. However, a move to Manchester United is projected to slash these outputs by up to 50% due to large stylistic differences between the teams more than differences in league difficulty, with de Jong no longer even ranking in the top 30% of central midfielders in the Premier League in each of these metrics.
- **Max Aarons to Bayern Munich (Right Back): Hot**
  Aarons has somewhat surprisingly been linked with Bayern Munich, Barcelona, and Manchester United considering he was playing in the EFL Championship last season. His high numbers of xA, final third passes, and take-ons—each in the top 15% of right backs in the EFL Championship—translate well to Bayern Munich, ranking in the top 30% of right backs in the Bundesliga, suggesting the 21-year-old may prove a shrewd investment.
- **Raheem Sterling to FC Barcelona (Winger): Tepid**
  Sterling has high production metrics compared to other wingers in the Premier League but relatively low numbers compared to other wingers at Manchester City. This pattern is predicted to carry over following a transfer to Barcelona, where he ranks in the 85th percentile of xG and 78th percentile of xA compared to wingers in La Liga—lower than most other Barcelona wingers. A good but possibly not great transfer considering the enormous fee being touted for his signature.
- **Karim Adeyemi to Borussia Dortmund (Striker): Hot**
  The 19-year-old Adeyemi has been exceptional for Salzburg, with his xA, xG and take-ons all in the top 20% of strikers in the Austrian Bundesliga. The transition to a much tougher league with Dortmund sees a predicted drop to the top 33% of strikers in the Bundesliga in terms of xG and xA, although this still shows great potential for a 19-year-old.
- **Aaron Mooy to Celtic (Central Midfielder): Hot**
  Despite a move from the Chinese Super League to the more competitive Scottish Premiership, Mooy's impressive versatility—in the top 20% of central midfielders for xG, xA, crosses, final third passes, and take-ons—is projected to hold up well. He would rank in the top 10% of central midfielders in the Scottish Premiership for xG and final third passes and in the top 20% of xA and take-ons, making him a prime candidate for Celtic manager and Australian compatriot Ange Postecoglou.
- **Rhys Healey to Brighton & Hove Albion (Striker): Tepid**
Healey is a very shooting-oriented player, ranking in the 98th percentile of strikers in Ligue 2 in terms of shots and xG per 90 but only around the 60th percentile in terms of passing and dribbling metrics. Whilst his shooting metrics are expected to fall by around 30% at Brighton—to that of an average striker in the Premier League—this may represent a relatively successful transfer for Brighton who often find themselves threatened with relegation and Healey is projected to be the club's most prolific striker.

## 6. Related Work and Summary

In this paper we have proposed a unique solution to identifying and predicting the performance of transfer targets in soccer, with the flexibility to be used across any team and league for which event level data is collected. Our goal was to predict transfer success by evaluating various player level per 90 metrics across multiple performance areas in attack and defence, whilst considering the ability and style of both player and teams. This differs from previous work predicting future player performance in sports which are less dependent on within-team player interactions, such as *PECOTA* in baseball [7], or aggregates to coarser outputs such as goals for and against that a soccer player provides for their team by Imburgio & Goldberg [8], or NBA team level season performance such as *FiveThirtyEight*'s roster-shuffling machine [9].

More dedicated player transfer prediction models are typically found in draft prediction, such as Pelton's wins above replacement player projections *WARP* in basketball [10]. In some cases, these models have started to utilize tracking data to identify more detailed player metrics such as pressure or lane blocking metrics in NBA draft projections such as Patton *et al.* [11]. For our Transfer Portal model to provide the required global coverage to identify prospects, we currently restrict the input data to event level.

A key contribution to our work is the combining of elements across player and team style to infer how style and ability combine across player, team, and league levels to impact future player performance. Previous work on style in soccer includes *SciSports'* [12] player role labels which assign players one of 22 pre-defined roles and the player2vec approach by Torvaney [13], which utilizes a natural language processing word2vec approach to provide 16-dimensional vector representations of players which can be compared to each other. There has also been work on provide more ability-focused measures, such as *SciSports* [12], *Analytics FC* [14] and *smarterscout*'s [15] single value rating system, which assigns ability scores to different facets of a player's game—and in the case of *smarterscout* is adjusted to the league quality. The Transfer Portal model aims to bring together these ideas of style and quality to help inform future performance.

Our feature pipeline and resulting deep learning Transfer Portal model creates a system which on average produces predictions of transfer performance with a 49% reduction in mean squared error when compared to using the raw player metrics obtained at their previous club. This highlights the value of creating a detailed understanding of not just the relative ability of teams and leagues involved in a transfer, but also the fit of a player based on player and team stylistic measures.

We have also shown the importance of this work through case studies in Section 4. We provided examples of how our model can support team management decision making when analyzing potential transfer targets, such as creating a shortlist of replacements for Doku at Stade Rennais FC and then investigating in detail their potential performance across a variety of different metrics. However, this is just scratching the surface of what the Transfer Portal model can provide to users. It can be used to determine whether current team players will retain their level of performance in the future, or answer a wide range of "what-if" questions such as: How might a target player perform if moved from a relegated threatened team to title candidate? How might target players perform at a club in a new country? How might they perform in a lower tier? How might a rising star perform following a breakout season? We touched on this using Transfer Portal for transfer rumours to quickly craft a narrative on the predicted success of the latest transfer rumours reported by the media.

Future work on Transfer Portal will expand the range of metrics available to predict on to include indirect defensive metrics—such as 'opposition allowed' metrics which quantify questions like 'how much less xG would Paris Saint-Germain concede after signing Sergio Ramos in defence?'—and goalkeeping metrics such as Expected Goals on Target (xGoT). We also look to move beyond positional labels like central defender, right back, and striker—which can ignore the diversity of playing styles amongst players who operate in similar areas of the pitch—to include Player Roles [16] [17].

Ultimately, transfer decisions depend on a multitude of different factors, from data analysis to scouting to interpersonal negotiations. Our Transfer Portal model utilises the power of deep learning models over thousands of historic transfers to help decision makers better inform vital recruitment choices which can make or break the success of their club.

# Appendix

## A. Adjustment Models

### A.1. Team Adjustment Model

The aim of this model is to adjust each team feature for the first game of a new league, based on the change of both team and league ability scores between the final game of their previous season and the first game on their new season.

For example, say we have a high Expected Goals (xG) team who get promoted, we might expect their xG per 90 in their first season in the new league of a higher tier to be much lower than in their promotion season. Therefore, we would want to adjust the initial xG per 90 feature value in their new league to one which is more reasonable given their new team and league ability.

To improve the initial team values, we train a so called "adjustment model" which predicts the feature value of the new season based on two pieces of information:

1. The distribution of the feature value in their new league. We expect that a newly promoted team will be in the lower percentiles of their new league for metrics like xG per 90, but higher percentiles for metrics like defensive actions. We will call this the naive expectation based on league information.

2. The team's feature value compared to the distribution in their previous league. For example, how does a promoted team compare to their previous league's 75% percentile in xG per 90, or the 25% percentile for defensive actions? We will call this the team's relative feature value in the previous league and is a measure of the team's relative ability in the previous league for each feature, to help us adjust our predictions from the naive expectation.

To train this model on rolling features we carefully construct target data based on the aims for this model. Since the data is rolling, we cannot use overlapping games, by which we mean separate games which might be used in the same rolling window as each other.

The aim of this model is to provide an initial value which is then ignored after a specific game or minute threshold is met, hence we consider the target to be predicting the team per 90 rolling features in the new league once this threshold is met. Say for example that the threshold is 2000 minutes before our team features ignore their prior values. Then the team adjustment model should be used to provide a reasonable approximation to how a team's features will change between the end of the previous season and 2000 minutes into their new league season.

To do this we define the targets as the team per 90 rolling values from the first game of their new season once the minutes threshold is met. Currently we use only two features as mentioned above, the naive expectation based on league information feature is used as an offset, whilst the team's relative feature value in previous league is used as a standard feature. This is outlined in the chart below.

This model is called an adjustment model due to the fact it's predicting the target using the naive league expectation values as an offset. It needs to be simple and consistent, which is why we decided on a regression model defined as

$$y_{i,j} = x_{i,j} + \alpha_j + \beta_j z_{i,j} + \epsilon_{i,j},$$

for targets $j = 0, \ldots, n$ and data points $i = 0, \ldots N,$ where

| Notation | Description |
|---|---|
| $y_{i,j}$ | Target value for the $i$th team, $j$th feature (team per 90 minute value after reaching new league minutes threshold). |
| $x_{i,j}$ | Naive expectation offset based on league information for the $i$th team, $j$th feature. |
| $z_{i,j}$ | Team's relative feature value in previous league for the $i$th team, $j$th feature. |
| $\epsilon_{i,j}$ | Independent and identically distributed error term (assumed Gaussian) for player $i$ and feature $j$. |

The $\alpha$ and $\beta$'s are our parameter estimates, which differ for each target.

Summarizing, we fit $n = 13$ independent models which predict each feature as a function of naive league assumption and previous season feature values.

### A.2. Team-Position Adjustments

We also require an adjustment of features on a finer level for use in the player feature adjustment model, where we record the team per 90 values aggregated by position. We do this by adjusting the team-position level features by the same percentage as the team level features were changed using the team adjustment regression model.

For example, say we are adjusting features for a newly promoted team using the team adjustment model, and the team xG per 90 is adjusted from 1.5 to 0.9. This is a reduction of 40%. Now let us assume that the xG per 90 of this team was 1.0 for strikers and 0.05 for centre backs prior to the adjustment. To adjust these team-position level xG per 90, we are also going to also reduce them by 40%. These predictions are highlighted in green in the table below

|  | xG per 90 minutes | Adjusted xG per 90 minutes | Percentage Change |
|---|---|---|---|
| Overall (full team using team adjustment model) | 1.5 | 0.9 | -40% |
| Strikers only | 1.0 | 0.6 | -40% |
| Centre Backs only | 0.05 | 0.03 | -40% |

### A.3. Player Adjustment Model

The aim of this model is to adjust each player feature for the first game of a new league, team, or position, based on previously known information about the player, team and league.

For example, if we have a player playing at Centre Back (CB) and their team is promoted, what do we consider a decent prior value for their features in the new league? Another example might be a CB joining a new team and we need a decent prior value for their features.

To improve the initial player values, we train another adjustment model which predicts the feature value of the new team/league/position based on four pieces of information:

1. The player's previous per 90 feature value for the player in the position we are predicting for, if available. This could be either in a previous team or league.

2. The average feature value for players in their position in the new team. We expect that a player will be somewhat like their teammates in the same position.

3. The difference in average feature value for players in their position between their old and new team. Again, this tells us how the teams play that the player is moving between. If they are just moving league within the same team, then this would be comparing the team's features in the previous league against our team feature projections in their new league.

4. The change in relative ability between the teams. We define relative ability as the difference between the team Power Ranking score and the league average score. Hence a positive valued team is better than the league average and negative valued team is worse than league average. For example, this feature can tell us whether the player is moving from a team doing well in their division to one doing badly. Again, if the player is moving leagues within the same team, then we compare how that team's relative ability changes between leagues.

The player feature adjustment model uses a linear regression model similar to that used in the team feature adjustment model. Our model is defined as

$$y_{i,j,k} = \alpha_j + \beta_{1,j} x_{1,i,j,k} + \beta_{2,j} x_{2,i,j,k} + \beta_{3,j} x_{3,i,j,k} + \beta_{4,j} x_{4,i,j} + \beta_{5,j} x_{4,i,j}^2 + \beta_{6,j} x_{4,i,j}^3 + \epsilon_{i,j,k},$$

for targets $j = 0, \ldots, n$, players $i = 0, \ldots N$, and positions $k = 0, \ldots, K$ where

| Notation | Description |
| --- | --- |
| $y_{i,j,k}$ | Target value for the $i$th player, $j$th feature in the $k$th position (player per 90 minute values after reaching minutes threshold). |
| $x_{1,i,j,k}$ | The player's previous per 90 minute feature value for the $i$th player, $j$th feature in the $k$th position. |
| $x_{2,i,j,k}$ | The average feature value for players in their position in the new team for the $i$th player, $j$th feature in the $k$th position. |
| $x_{3,i,j,k}$ | The difference in average feature value for players in their position between their old and new team for the $i$th player, $j$th feature in the $k$th position. |
| $x_{4,i,j}$ | The change in relative ability between the teams for the $i$th player, $j$th feature. |
| $\epsilon_{i,j,k}$ | Independent and identically distributed error term (assumed Gaussian) for player $i$, feature $j$ and position $k$ |